\documentclass{article}
\usepackage{graphicx} 
\usepackage{mystyle}

\title{Attention-Only Transformers and Implementing MLPs with Attention Heads}
\author{Robert Huben and Valerie Morris}
\date{}

\begin{document}

\maketitle

\begin{abstract}
    The transformer architecture is widely used in machine learning models and consists of two alternating sublayers: attention heads and MLPs. We prove that an MLP neuron can be implemented by a masked attention head with internal dimension 1 so long as the MLP's activation function comes from a restricted class including SiLU and close approximations of ReLU and GeLU. This allows one to convert an MLP-and-attention transformer into an attention-only transformer at the cost of greatly increasing the number of attention heads. We also prove that attention heads can perform the components of an MLP (linear transformations and activation functions) separately. Finally, we prove that attention heads can encode arbitrary masking patterns in their weight matrices to within arbitrarily small error. 
\end{abstract}

\section{Introduction}

The transformer architecture 
was introduced in the landmark 2017 paper \emph{Attention is All You Need} \protect\citep{vaswani2023attention} and traditionally consists of alternating attention and multilayer-perceptron (MLP) sublayers. Although initially used for machine translation, transformers have been used across a wide range of tasks, including language modeling \citep{radford2018improving, devlin2019bert, liu2018generating}, computer vision \citep{10.1145/3505244, Cornia_2020_CVPR}, and image generation \citep{pmlr-v80-parmar18a}. The widespread deployment of transformers has led to increasing interest in \emph{mechanistic interpretability} \citep{wang2022interpretability, conmy2023towards}, which seeks to convert the computations of transformers into human-understandable explanations. Some interpretability efforts, such as \citet{elhage2021mathematical}, focused on attention-only transformers, finding that MLP layers were harder to interpret.

This work seeks to supplement those mechanistic interpretability methods by showing that MLP layers in transformers are equivalent to a sum of masked attention heads and therefore can be subjected to interpretability techniques that work on attention-only transformers. In Theorem \ref{thm:mlp-attention-equality} we show that by including a ``bias token'' akin to the persistent memory vectors in \citet{sukhbaatar2019augmenting} and using a slightly unusual attention-masking pattern, an MLP layer of size $\ell$ can be written as the sum of $\ell$ attention heads with internal dimension 1. We show in Theorem \ref{thm:attention_only} that one can apply this process throughout the entire transformer, converting the typical MLP-and-attention transformer into an attention-only transformer. We then show in Theorems \ref{thm:linear_via_attention} and \ref{thm:activation_via_attention} that attention heads can implement row-wise linear transformations and matrix-level activation functions separately. Finally, we show in Theorem \ref{thm:pseudo_masking} that a slightly augmented network is capable of approximating any masking pattern to within arbitrary error.

\section{Background}
\begin{notation*}
    Throughout, we will use $M_{n,k}$ to denote the set of real-valued $n$-by-$k$ matrices. 
    
    For matrices $X \in M_{n_1, k_1}$ and $Y \in M_{n_2, k_2}$ of any size, we will write $X \oplus Y$ for the block matrix
    
    \[ X \oplus Y =\left[ \begin{array}{c|c}
          X & \mathbf{0} \\
          \hline
          \mathbf{0} & Y 
        \end{array} \right] \in M_{n_1+n_2, k_1+k_2} \]
    
    where each $\mathbf 0$ is a correctly sized zero matrices. We will similarly write $\mathbf{1}$ for matrices with a 1 for every entry.

    For matrices $X \in M_{n, k_1}$ and $Y \in M_{n, k_2}$, we will write 
    
    \[ [X|Y] \in M_{n, k_1+k_2}\] for the matrix made by appending one to the other. 

    For a real-valued function $f$ and matrix $X$, we will write $f(X)$ for the entry-wise application of that function to the matrix. 
    
    We write   
    \begin{eqnarray*}
        \mathrm{ReLU}(x):=& max(x,0) \\
        \mathrm{SiLU}(x):=& x\sigma(x)\\
        \mathrm{GeLU}(x):=& x \Phi(x)\\
    \end{eqnarray*}

    where $\sigma(x)=1/(1+\exp(-x))$, and $\Phi(x)$ is the cumulative distribution function for the standard Gaussian distribution with mean 0 and variance 1. We will say that a \emph{generalized SiLU function} is a function of the form 
    
    \[ f(x)=a_1 \mathrm{SiLU}(a_2x)\] 
    
    for some $a_1,a_2 \in \R$.

\end{notation*}

The class of generalized SiLU functions includes $\mathrm{SiLU}(x)$ and approximations of $\mathrm{GeLU}$ and $\mathrm{ReLU}$. In particular, $\mathrm{GeLU} (x) \approx \mathrm{SiLU}(1.702x)/1.702$ \citep{hendrycks2023gaussian} (reaching a maximum absolute error of $0.0203$ at $x= \pm 2.27$) and $\mathrm{ReLU}(x) \approx \mathrm{SiLU}(k x)/k$ for large $k$ (reaching a maximum absolute error of $\frac {0.2785}k$ at $x=\pm \frac{1.278}k$).


\begin{definition}
An \emph{MLP with no biases and one hidden layer} is a function $f:M_{n,k} \ra M_{n,k}$ of the form

\begin{equation}
    f(X)=\label{eqn:mlp}
    \alpha(XV_1)V_2
\end{equation} 

where $\alpha:\R \ra \R$ is some real-valued function applied entry-wise to matrices, and $V_1, V_2$ are fixed matrices in $M_{k, \ell}$ and $M_{\ell,k}$, respectively, called parameter matrices. The number $\ell$ is called the \emph{size of the hidden layer}, and the function $\alpha$ is called the \emph{activation function}. 

\end{definition}

Many transformer architectures follow the convention that $\ell=4k$ \citep{vaswani2023attention,brown2020language}, but we do not require this. There are many popular choices for activation functions \citep{hendrycks2023gaussian}, including $\mathrm{ReLU}$, $\mathrm{SiLU}$, and $\mathrm{GeLU}$.

For describing attention heads, we largely follow the framework of \citet{elhage2021mathematical}.

\begin{definition}
    A \emph{mask matrix} $\Lambda$ is a matrix with entries in $\{0,1\}$ such that every row has at least one nonzero entry.

    Let $X, \Lambda \in M_{n,k}$, and suppose $\Lambda$ is a mask matrix. Then define the \emph{masked softmax} function 
    
    \[ \msoftmax(X, \Lambda):= \mathrm{rownorm}\left(\exp(X) \odot \Lambda \right)\]
    
    where $\mathrm{rownorm}$ denotes row-wise $\ell^1$ normalization, and $\odot$ denotes element-wise multiplication. That is, the masked softmax function acts like the usual row-wise softmax but applied to only the entries of $X$ where the mask $\Lambda$ is 1. At the entries where $\Lambda$ is 0, the output of the masked softmax function takes the value 0.

    A \emph{masked attention head} is a function $h:M_{n,k} \ra M_{n,k}$ of the form

    \begin{equation}
    \label{eqn:attention_head}
    h(X)= \msoftmax(XW_{QK}X^T, \Lambda) X W_{OV}        
    \end{equation} 
    
    for some matrices $W_{OV}, W_{QK} \in M_{k,k}$, and mask matrix $\Lambda \in M_{n,n}$. We call $W_{OV}$ and $W_{QK}$ the \emph{parameter matrices} for this attention head.
\end{definition}

For practical reasons, attention heads are rarely described (or implemented) as in Equation \ref{eqn:attention_head}. However, one can verify that this definition encompasses the classical transformer framework 
in \citet{vaswani2023attention}, with $W_{QK}=(W^Q_i)(W^K_i)^T/\sqrt{d_k}$, and $W_{OV}=W^V_iW^O_i$, where $W^O_i$ denotes the appropriate subblock of the $W^O$ matrix.

For many language tasks, the masking pattern is chosen to mask later tokens from earlier tokens \citep{vaswani2023attention, radford2018improving}, i.e., $\Lambda$ is the subdiagonal matrix with $\Lambda_{i,j} = \begin{cases} 1 & \text{ if } i \leq j \\ 0 & \text{ otherwise} \end{cases}$. However, in our construction in Theorem \ref{thm:mlp-attention-equality}  and Theorem \ref{thm:attention_only}, we will make use of a nonstandard masking pattern in which tokens only attend to themselves and a single special token.

\section{Implementing MLP Layers with Attention Heads}

In this section we show that MLP layers whose activation functions are generalized SiLU functions are in fact a sum of attention heads. 

The intuition for this claim is simple: both attention heads and MLPs are mostly linear, with a single nonlinearity (respectively, masked softmax and the generalized SiLU activation function). Additionally, softmax can easily play the role of the sigmoid part of SiLU since $\mathrm{softmax}([-x,0])=\mathrm{rownorm}([e^{-x}, 1])= [\sigma(x), \sigma(-x)]$. Multiplying this attention pattern onto the vector $[x,0]$, we get $x\sigma(x)+0\sigma(-x)=\mathrm{SiLU}(x)$. The following theorem is a formalization of this intutition.

\begin{theorem}
\label{thm:mlp-attention-equality}
    Let $f(X)=\alpha(XV_1)V_2$ be an MLP on $M_{N,D}$ with no biases and one hidden layer of size $\ell$, and suppose $\alpha$ is a generalized SiLU function $\alpha(x)=a_1 \mathrm{SiLU}(a_2x)$. Then there are $\ell$ masked attention heads $\{h_i\}_{i=1}^\ell$ on $M_{N+1, D+1}$ such that

    \[ f(X) \oplus [0]=\fsum_{i=1}^\ell h_i(X \oplus [1])\]

    for all $X \in M_{N,D}$.

    In particular, for the $i$th attention head, one uses parameter and mask matrices 

    \begin{eqnarray*}
    W_{QK} &=& a_2 \left[ \begin{array}{c|c}
      \mathbf{0} & -V_1^i \\
      \hline
      \mathbf{0} & 0
    \end{array} \right]\\
    W_{OV} &=&a_1a_2V_1^iV_2^i \oplus [0]\\
    \Lambda &=& \left[ \begin{array}{c|c}

      I_N & \mathbf{1} \\
      \hline
      \mathbf{0} & 1
    \end{array} \right]
    \end{eqnarray*} 
    
    where the block decompositions are into size $N$ and $1$, $V_1^i$ denotes the $i$th column of $V_1$, $V_2^i$ denotes the $i$th row of $V_2$, and $\mathbf{1}$ denotes the column vector of all 1s.
\end{theorem}

\begin{proof}
We first prove the claim in the case of $\ell= a_1=a_2=1$. In this case, since there is only one column in $V_1$, then $V_1=V_1^i$, and similarly $V_2=V_2^i$. Consider the attention matrix $\msoftmax((X \oplus [1])W_{QK}(X \oplus [1])^T, \Lambda)$. Multiplying matrices on the level of their blocks, we get that the first argument of the masked softmax is

\[ (X \oplus [1])W_{QK}(X \oplus [1])^T= \left[ \begin{array}{c|c}
      X & \mathbf{0} \\
      \hline
      \mathbf{0} & 1 
    \end{array} \right] \left[ \begin{array}{c|c}
      \mathbf{0} & -V_1^i \\
      \hline
      \mathbf{0} & 0
    \end{array} \right]\left[ \begin{array}{c|c}
      X & \mathbf{0} \\
      \hline
      \mathbf{0} & 1 
    \end{array} \right]^T = \left[ \begin{array}{c|c}
      \mathbf{0} & -XV_1 \\
      \hline
      \mathbf{0} & 0
    \end{array} \right]\]

Now consider the masked softmax term in the $j$th row for $j\leq N$. This row has exactly two unmasked values, the diagonal entry and the rightmost entry, taking the values 0 and $-(XV_1)_j$, respectively. Applying $\exp$ and $\mathrm{rownorm}$ results in $\sigma((XV_1)_j)$ and $\sigma(-(XV_1)_j)$, respectively. Thus, the masked softmax term becomes

\begin{eqnarray*}
    \msoftmax((X \oplus [1])W_{QK}(X \oplus [1])^T, \Lambda) &=& \msoftmax(\left[ \begin{array}{c|c}
      \mathbf{0} & -XV_1 \\
      \hline
      \mathbf{0} & 0
    \end{array} \right], \left[ \begin{array}{c|c}
      I_{n-1} & \mathbf{1} \\
      \hline
      \mathbf{0} & 1
    \end{array} \right]) \\
    &=& \left[ \begin{array}{c|c}
      \mathrm{diag}(\sigma(XV_1)) & \sigma(-XV_1) \\
      \hline
      \mathbf{0} & 1 
    \end{array} \right]
\end{eqnarray*}

Substituting these values into the expression for $h(X)$ gives

\begin{eqnarray*}
    h(X \oplus [1]) &=& \msoftmax((X \oplus [1])W_{QK}(X \oplus [1])^T, \Lambda) (X \oplus [1]) W_{OV} \\
    &=& \left[ \begin{array}{c|c}
      \mathrm{diag}(\sigma(XV_1)) & \sigma(-XV_1) \\
      \hline
      \mathbf{0} & 1 
    \end{array} \right] (X \oplus [1]) W_{OV} \\
    &=& \left[ \begin{array}{c|c}
      \mathrm{diag}(\sigma(XV_1)) & \sigma(-XV_1) \\
      \hline
      \mathbf{0} & 1 
    \end{array} \right]
     \left[ \begin{array}{c|c}
      X & \mathbf{0} \\
      \hline
      \mathbf{0} & 1 
    \end{array} \right]
     \left[ \begin{array}{c|c}
      V_1V_2 & \mathbf{0} \\
      \hline
      \mathbf{0} & 0 
    \end{array} \right]\\
    &=& \left[ \begin{array}{c|c}
      \mathrm{diag}(\sigma(XV_1))XV_1V_2 & \mathbf{0} \\
      \hline
      \mathbf{0} & 0 \end{array} \right]\\
    &=& \left[ \begin{array}{c|c}
      \mathrm{SiLU}(XV_1)V_2 & \mathbf{0} \\
      \hline
      \mathbf{0} & 0 \end{array} \right]\\
    &=& \left[ \begin{array}{c|c}
      f(X) & \mathbf{0} \\
      \hline
      \mathbf{0} & 0 \end{array} \right]\\
    &=& f(X) \oplus [0]
\end{eqnarray*}

as desired. This completes the $\ell=a_1=a_2=1$ case.

For a general $a_1, a_2$, apply the previous case to an MLP with weight matrices $a_2V_1$ and $a_1V_2$.

Finally, for the fully general case with $\ell>1$, for each $1 \leq i \leq \ell$, let $f_i(X)=\alpha(XV_1^i)V_2^i$, and note that $f=\fsum_{i=1}^\ell f_i$. Let $h_i$ denote the attention head corresponding to $f_i$ given by the $\ell=1$ case. Then we have that

\begin{eqnarray*}
    f(X) \oplus [0] &=&\fsum_{i=1}^\ell f_i(X) \oplus [0] \\
    &=&\fsum_{i=1}^\ell h_i(X \oplus [1])
\end{eqnarray*}
as desired.
\end{proof}

\begin{remark}
The additional term $\oplus [1]$ in Theorem \ref{thm:mlp-attention-equality} is similar to the \emph{persistent vectors} of \citet{sukhbaatar2019augmenting}. In that work, the authors propose a new architecture, which they call the all-attention architecture, in which attention can also be paid to certain static vectors, learned for each attention head, called the persistent vectors. Our approach could also be implemented in that architecture with a single persistent vector $(0,0,0,..,0,1)$ shared across all attention heads.

Note also that the $W_{QK}$ and $W_{OV}$ matrices used in Theorem \ref{thm:mlp-attention-equality} can be factored into the matrices $W_Q$, $W_K$, $W_V$, $W_O \in M_{D+1, 1}$ from \citet{vaswani2023attention} satisfying $W_{QK}=W_QW_K^T/\sqrt{D+1}$ and $W_{OV}=W_VW_O$. In particular, we can take $W_Q=W_V=a_2[V_1^i | 0]^T$, $W_K=\sqrt{D+1}[\mathbf{0}|-1]^T$, and $W_O=a_1[V_2^i|0]^T$. Since $W_K$ is shared across all attention heads, we only need to store two sets of parameters, the vectors $W_Q=W_V$ and $W_O$. 

This provides an alternative perspective on MLP neurons: a neuron in an MLP is an attention head with internal dimension 1 and a particularly restrictive masking pattern in which each token attends only to itself and a static ``bias'' token.
\end{remark}

We now have the necessary tools to show that a decoder-only transformer as in \citet{liu2018generating, radford2018improving} can be implemented entirely with attention heads.

\begin{definition}
A transformer is a function $t:M_{N,D} \ra M_{N,D}$ of the form $X_0 \mapsto X_1 \mapsto... \mapsto X_m = t(X_0)$, where

\begin{equation*}
    X_{j+1} = \begin{cases}
        \mathrm{\mathrm{LayerNorm}}(X_{j}+ \fsum_i h_{j, i}(X_{j})) & \text{ or} \\
        \mathrm{\mathrm{LayerNorm}}(X_{j}+ f_j(X_{j})) &
    \end{cases} 
\end{equation*}

for some attention heads $h_{j,i}$ or MLPs with a single hidden layer $f_{j}$. Note the use of Layer Normalization \citep{ba2016layer} and \emph{skip connections}, where one performs some computation $f$ on $X_j$ and defines $X_{j+1} = \mathrm{LayerNorm}(X_j+f(X_j))$, as opposed to $X_{j+1}=f(X_j)$.
\end{definition}

Classically, transformers alternate between attention sublayers and MLP sublayers, but we allow the existence of other architectures, including attention-only transformers and ``MLP-only'' transformers.

\begin{theorem}
    \label{thm:attention_only}
    If a transformer's MLP layers are activated by a generalized SiLU function, they can be substituted with attention heads.
\end{theorem}

\begin{proof}
We will show that we can create a new transformer $t'$ on $M_{N+1, D+1}$ whose residual stream $X_j'$ on every sublayer satisfies

\[ X_j' =X_j \oplus [1]\]

This is sufficient to prove the main claim since the output of this new transformer will be $X_{2m}'=X_{2m}\oplus [1]$ and therefore contain the output of the original transformer.

Without loss of generality, assume that the MLP layers have no bias terms (i.e., that we've already used the ``bias trick'' to fold bias terms into the weight matrix).

To prove that there is a transformer $t'$ that satisfies $X_j'=X_j \oplus [1]$ on every sublayer, we proceed by induction. For the base case of $j=0$, we tweak the transformer's context window and embedding weights so that $X_0'=X_0 \oplus [1]$.

We split the inductive case depending on whether the original transformer's sublayer used attention or an MLP. If the original layer was an MLP, then by Theorem \ref{thm:mlp-attention-equality}  there are attention heads $h_{j,i}'$ such that $f_j(X) \oplus [0]= \fsum h_{j,i}'(X \oplus [1])$, so in our transformer $t'$, using these attention heads yields 

\begin{eqnarray*}
    X_{j+1}' &=& \mathrm{LayerNorm} (X_j' + \fsum h_{j,i}'(X_j')) \\
    &=& \mathrm{LayerNorm} ((X_j \oplus [1]) + \fsum h_{j,i}'(X_j \oplus [1])) \\
    &=& \mathrm{LayerNorm} ((X_j \oplus [1]) + (f_j(X) \oplus [0]))) \\
    &=& \mathrm{LayerNorm} (X_j + f_j(X)) \oplus [1] \\
    &=& X_{j+1} \oplus [1]
\end{eqnarray*}

as desired.

If instead, the transformer used attention heads on the $j$th sublayer, we must tweak our original induction heads to account for the new size. To this end, we will show that for each of the original induction heads $h=h_{j,i}$, we can create an induction head $h'$ such that

\[ h'(X \oplus [1]) = h(X) \oplus [0]\]

Let $W_{QK}, W_{OV}$, and $\Lambda$ denote the original parameter and masking matrices for $h$. Then define

\begin{eqnarray*}
W_{QK}' &=& W_{QK} \oplus [1]\\
W_{OV}' &=& W_{OV} \oplus [0]\\
\Lambda' &=& \Lambda \oplus [1]
\end{eqnarray*} 

Then,

\begin{eqnarray*}
    h'(X \oplus [1]) &=& \msoftmax((X \oplus [1])W_{QK}' (X \oplus [1])^T, \Lambda') (X \oplus [1]) W_{OV}' \\
    &=& \msoftmax((X \oplus [1])(W_{QK} \oplus [1]) (X \oplus [1])^T, (\Lambda \oplus [1])) (X \oplus [1]) (W_{OV} \oplus [0]) \\
    &=& \msoftmax(XW_{QK}X^T \oplus [1], \Lambda \oplus [1]) (XW_{OV} \oplus [0])\\
    &=& (\msoftmax(XW_{QK}X^T, \Lambda) \oplus [1]) (XW_{OV} \oplus [0]) \\
    &=& \msoftmax(XW_{QK}X^T, \Lambda)XW_{OV} \oplus [0] \\
    &=& h(X) \oplus [0]
\end{eqnarray*}

as desired. Now, creating such $h'_{j,i}$ for each of the original attention heads $h_{j,i}$, we have

\begin{eqnarray*}
    X_{j+1}' &=& \mathrm{LayerNorm} (X_j' + \fsum h_{j,i}'(X_j')) \\
    &=& \mathrm{LayerNorm} ((X_j \oplus [1]) + \fsum h_{j,i}'(X_j \oplus [1])) \\
    &=& \mathrm{LayerNorm} ((X_j \oplus [1]) + \fsum h_{j,i}(X) \oplus [0])) \\
    &=& \mathrm{LayerNorm} ((X_j + \fsum h_{j,i}(X))) \oplus [1] \\
    &=& X_{j+1} \oplus [1]
\end{eqnarray*}

as desired. This completes the inductive step and the proof.

\end{proof}

It is instructive to compare this construction to the negative results of \citet{dong2021attention}, which find that without skip connections or MLPs, a self-attention network converges rapidly to a rank-1 matrix. Since we obviously do away with the MLP layer, our result depends on the use of skip connections. In particular, the ``bias term'' of $\oplus [1]$ is zeroed out by the construction in Theorem \ref{thm:mlp-attention-equality}, so applying the construction in Theorem \ref{thm:attention_only} without a skip connection results in $X_0'=X_0 \oplus [1]$, but $X_1'=X_1 \oplus [0]$. Then, in the $j=2$ sublayer, the construction in \ref{thm:mlp-attention-equality} would fail for lack of this bias term, as, without it, the pre-attention matrix $(X')W_{QK}(X')^T$ is 0.




\section{Linear Transformations and Activation Functions with Attention Heads}

Theorem \ref{thm:mlp-attention-equality}  shows that attention heads can implement an MLP layer, but can they separately implement the components of an MLP, a linear transformation and an activation function? In this section we show that the answer is yes.

We first show that an attention head can perform an arbitrary linear operation row-wise on the matrix.

\begin{theorem}
    \label{thm:linear_via_attention}
    Let $h:M_{N,D} \ra M_{N,D}$ be an attention head with masking matrix $\Lambda=I_N$. Then $h(X)=XW_{OV}$.
\end{theorem}

\begin{proof}
    Because $\Lambda=I_n$, after masking, the attention matrix $\msoftmax(XW_{QK}X^T, \Lambda)$ will have nonzero entries only along the diagonal. Since the rows of the attention matrix are normalized to sum to 1, it follows that $\msoftmax(XW_{QK}X^T, \Lambda)=I_n$. Then, 
    \[ h(X)= \msoftmax(XW_{QK}X^T, \Lambda) X W_{OV} =I_n X W_{OV}=XW_{OV}\]
    as desired.
\end{proof}

Now we will show that one can apply a generalized SiLU function entrywise.

\begin{theorem}
    \label{thm:activation_via_attention}
    Let $\alpha$ be a generalized SiLU function. Then there are $D$ attention heads $h_1,..., h_D$ on $M_{N+1,D+1}$ such that 
    
    \[ \alpha(X) \oplus [0]= \fsum_{i=1}^D h_i(X \oplus [1])\]

\end{theorem}

\begin{proof}
    This follows immediately from applying Theorem \ref{thm:mlp-attention-equality}  to the MLP $f(X)=\alpha(XI_N)I_N=\alpha(X)$, whose hidden layer is of size $\ell=D$.
\end{proof}

Note that a transformer usually makes use of skip connections, so that the residual stream experiences the transformation $X \mapsto X + sublayer(X)$. Thus, to get the transformation $X \mapsto \alpha(X)$, one can combine these two theorems, using $D+1$ attention heads to produce $sublayer(X)=\alpha(X)-X$, in which case $X \mapsto X+ sublayer(X) = \alpha(X)$. 

\section{Encoding Masking Patterns in Weight Matrices}

Although some previous work has used multiple masking patterns\footnote{E.g., \citet{brown2020language} uses ``alternating dense and locally banded sparse attention patterns''.}, some readers may be disappointed that the attention patterns prescribed in the previous sections are oddly ``artificial''. In this section, we will show a technique to ameliorate this concern by embedding the masking pattern into the $W_{QK}$ matrix. To do so, we must further augment the residual stream, but our technique allows us to encode an arbitrary masking pattern in the $W_{QK}$ parameters at the cost of arbitrarily small errors and poor training behavior.

\begin{theorem}
    \label{thm:pseudo_masking}
    Let $h$ be a masked attention head on $M_{N,D}$ with mask matrix $\Lambda_1$. Then for any mask matrix $\Lambda_2$ satisfying $\Lambda_1 \leq \Lambda_2$ entrywise, there is a family of masked attention heads $h_\Omega$, parameterized by $\Omega \in \R$, that use $\Lambda_2$ as their mask matrix and such that $h_\Omega([X|I_N]) \rightarrow [h(X)|\mathbf{0}]$ uniformly on compacta as $\Omega \ra \infty$.
\end{theorem}

\begin{proof}
    Define $h_\Omega$ to be the attention head using the mask matrix $\Lambda_2$ and parameter matrices 
    
    \begin{eqnarray*}
    W_{QK, \Omega} &=& W_{QK} \oplus \Omega \Lambda_1 \\
    W_{OV, \Omega} &=& W_{OV} \oplus \mathbf 0
    \end{eqnarray*}
    
    Fix some compact set $K \subset M_{N,D}$ and $\epsilon>0$.
    
    First observe that

    \begin{eqnarray*}
        h_\Omega([X|I_N]) &:=& \msoftmax([X|I_N] W_{QK, \Omega} [X|I_N]^T, \Lambda_2) [X|I_N]W_{OV, \Omega}\\
        &=& \msoftmax([X|I_N] (W_{QK} \oplus \Omega \Lambda_1)[X|I_N]^T, \Lambda') [X|I_N](W_{OV} \oplus \mathbf 0)\\
        &=& \msoftmax(XW_{QK}X^T+ \Omega \Lambda_1, \Lambda_2) [XW_{OV}|\mathbf{0}]\\
    \end{eqnarray*}

    Our first task is to show that the attention pattern $A_1:=\msoftmax(XW_{QK}X^T+ \Omega \Lambda_1, \Lambda_2)$ converges to the corresponding attention pattern $A_2:=\msoftmax(XW_{QK}X^T, \Lambda_1)$ entrywise as $\Omega \ra \infty$. To this end, fix $\epsilon_0>0$, and pick $b \in \R$ such that entries of $XW_{QK}X^T$ are bounded in absolute value by $b$ as $X$ ranges over $K$, and let $\Omega> \ln(N/\epsilon_0)+2b$. We have three cases depending on whether the corresponding entries in $\Lambda_1$ and $\Lambda_2$ are $0$ or $1$:
    \begin{enumerate}
        \item If $\Lambda_{1, (i,j)}=\Lambda_{2 (i,j)}=0$, then $A_{1, (i,j)}=A_{2,(i,j)}=0$ due to masking.
        \item If $\Lambda_{1, (i,j)}=0$ and $\Lambda_{2, (i,j)}=1$, then $A_{1, (i,j)}=0$. Since $\Lambda_1$ is a mask matrix, in row $i$ there is a column $J$ such that $\Lambda_{1, (i,J)} =1$. Then the $(i,J)$th entry of $\exp(XW_{QK}X^T+ \Omega \Lambda_1)$ is at least $\exp(\Omega-b)$, while the $(i,j)$th entry is at most $\exp(b)$. Thus, after row-normalizing, we have
        
        \begin{eqnarray*}
            A_{2, (i,j)} &\leq& \frac{\exp(b)}{\exp(\Omega-b)} \\
            &=& \frac{1}{\exp(\Omega-2b)}
        \end{eqnarray*}
        
        Since $\Omega> \ln(N/\epsilon_0)+2b$, we have $\exp(\Omega-2b)>N/\epsilon_0$, so $A_{2, (i,j)} \leq \frac{1}{N/\epsilon_0}=\epsilon_0/N<\epsilon_0$ as desired.
        
        \item If $\Lambda_{1, (i,j)}=\Lambda_{2, (i,j)}=1$, then consider the $i$th row. As shown in the previous two cases, in each entry of this row where $\Lambda_{1, (i,j)}=0$, we have $A_{2, (i,j)}< \epsilon_0/N$. Since there are $N$ terms in this row, and any row sums to 1 due to normalization, this means that the remaining terms, where $\Lambda_{1, (i,j)}=1$, sum to some value $S \in [1-\epsilon_0, 1]$. Since the log ratio between two such terms is the difference of their corresponding entries in $XW_{QK}X^T+ \Omega \Lambda_1$, and the $\Omega$ terms of those entries will cancel, this shows that the ratio between terms where $\Lambda_{1, (i,j)}=1$ in $A_2$ is the same as the corresponding ratio in $A_1$. That is, the $i$th row of $A_1$ concentrates its mass $S$ in the same locations as $A_2$ at the same ratios, so $A_{1,(i,j)}=SA_{2,(i,j)}$ for all $j$ with $\Lambda_{1,(i,j)}=1$. Thus $|A_{1,(i,j)}-A_{2,(i,j)}| = A_{1,(i,j)}|1-S|<\epsilon_0$.
    \end{enumerate}

    Rephrasing our partial result, we have shown that $A_1=A_2+E_\Omega$, where $E_\Omega$ is an error matrix whose entries are bound by $\epsilon_0$ whenever $\Omega> \ln(N/\epsilon_0)+2b$. 
    
    Returning to our expression for $h_\Omega([X|I_N])$, we have

    \begin{eqnarray*}
        h_\Omega([X|I_N]) &=&A_1[XW_{OV}|\mathbf{0}]\\
        &=&  (A_2 +E_\Omega) [XW_{OV}|\mathbf{0}]\\
        &=&  A_1[XW_{OV}|\mathbf{0}] +E_\Omega[XW_{OV}|\mathbf{0}]\\
        &=&  [h(X)|\mathbf 0 ] +[E_\Omega XW_{OV}|\mathbf{0}]
    \end{eqnarray*}

    Thus, the entry-wise difference between $h_\Omega([X|I_N])$ and $[h(X)|\mathbf 0 ]$ is  $[E_\Omega XW_{OV}|\mathbf{0}]$, so it suffices to show that $E_\Omega XW_{OV}$ is entry-wise less than $\epsilon$. To this end, fixing some $\epsilon>0$, let $\epsilon_0=\epsilon/K$, where $K=\max(||XW_{OV}||/\sqrt{N}, 1)$ and $||\cdot||$ denotes the operator norm of a matrix. Then, for all $\Omega>\ln(N/\epsilon_0)+2b$, we have $E_\Omega$ is entry-wise less than $\epsilon_0$. Therefore, in the $i,j$th entry of $E_\Omega XW_{OV}$, we have
    
    \begin{eqnarray*}
    |(E_\Omega XW_{OV})_{i,j}| &=& |row_i(E_\Omega) \cdot column_j(XW_{OV})| \\
    &\leq & \epsilon_0 \sqrt N \cdot ||XW_{OV}|| \\
    & = & (\epsilon/K) \sqrt N||XW_{OV}|| \\
    & \leq & (\epsilon/(||XW_{OV}||/\sqrt{N})) \sqrt N||XW_{OV}||  \\
    & = & \epsilon
    \end{eqnarray*}

    as desired.
\end{proof}

The above result shows that by augmenting the residual stream with an $I_N$ matrix, one can write the masking pattern into the $W_{QK}$ matrix. Combined with Theorem \ref{thm:attention_only}, this shows that one can convert a standard transformer into one using only attention heads and the standard masking pattern.

\begin{remark}
    Inspecting the relation between $\epsilon$ and $\Omega$ in the previous theorem allows us to provide a more concrete choice of $\Omega$. We require $\Omega>\ln(N/\epsilon_0)+2b$, where $N$ is the size of the context window, $\epsilon_0=\epsilon/\max(||XW_{OV}||/\sqrt{N}, 1)$, and $b$ is a bound on the entries of $XW_{QK}X^T$. 

    Using properties of logs, we may simplify our requirement to

    \[ \Omega> \ln(N/\epsilon)+ 2b+\max(\ln(N^{\frac 1 2} ||XW_{OV}||), 0)\]

    Since the entries of a marix are bounded by the matrix's operator norm, we can take $b= ||XW_{QK}X^T||=||X||^2 ||W_{QK}||$. The resulting requirement on $\Omega$ is then an increasing function of $||X||$, so we may remove our dependence on it by replacing it with $B=\sup_{X \in K} ||X||$, in which case our bound becomes

    \[ \Omega> \ln(N/\epsilon)+ 2B^2 ||W_{QK}||+\max(\ln(N^{\frac 1 2} B||W_{OV}||), 0) \]

    Notably, $\Omega$ grows only in the logarithm of $\epsilon$.
\end{remark}

\begin{example}
    Let's compute a value of $\Omega$ that is suitable for a particular language model. Take $\epsilon=2^{-146}$, the minimum positive value representable by a single-precision floating-point number \citep{4610935}, and apply this to GPT-2, which has a maximum context window of $N=1024$ tokens \citep{radford2019language}. According to \citet{millidge2023basic}, individual model weights are normally distributed, falling entirely within $[-1,1]$. Recall that $W_{QK}$ is in fact stored internally as two matrices $W_Q$ and $W_K$, with $W_{QK}=W_QW_K^T$. Such matrices are conventionally of size $N \times D/n_{heads}$, and since $D=1600$ \citep{radford2019language}, and $n_{heads}=25$ \citep{heimersheim2023residual}, we have $W_Q, W_K \in M_{1024, 64}$. Combining this with the bound that each entry is in $[-1,1]$, we get that $||W_Q|| \leq \sqrt{64} =8$. Similarly, $||W_K|| \leq 8$, so $||W_{QK}|| \leq ||W_Q||||W_K|| \leq 8 \cdot 8 =64$. By a similar argument, $||W_{OV}|| \leq 64$.
    
    For the bound $B$ on the norm of the residual stream, we turn to \citet{heimersheim2023residual} who finds that the measured norm of the residual stream increases across layers but does not seem to exceed $B=10^4$. Combining these into our formula, we find that a sufficient value of $\Omega$ is 

    \begin{eqnarray*}
        \Omega &=& \ln(N/\epsilon)+ 2B^2 ||W_{QK}||+\max(\ln(N^{\frac 1 2} B||W_{OV}||_2), 0) \\
        &=& \ln(1024/2^{-146})+ 2(10^4)^2 \cdot 8+\max(\ln(1024^{\frac 1 2} 10^4 \cdot 8), 0) \\
        &\approx& 1.6\times 10^9
    \end{eqnarray*}

with almost all of the contribution due to the $2B^2 ||W_{QK}||$ term.

\end{example}

\section{Limitations}

The technique described in Theorem \ref{thm:attention_only} faces several practical limitations. First is the quantity of attention heads: we use one attention head per dimension of the hidden layer, which can easily increase the number of attention heads by several orders of magnitude, partially offset by the new attention heads having smaller internal dimension. For example, each layer of GPT-3 has 96 attention heads with internal dimension 128 \citep{brown2020language}, and the process we describe would require 49152 additional 1-dimensional attention heads in each layer.

Second, it may be the case that replacing a feedforward network with attention heads slows down model inference or training. In particular, this approach replaces matrix multiplication with many vector-by-vector multiplications. One also computes many terms that are ``thrown away'' in the masking step. Combined, these suggest that converting an MLP layer to attention heads would increase computational costs.

Finally, the ``pseudo-masking'' in Theorem \ref{thm:pseudo_masking} introduces a separate set of issues into any training process due to the large $\Omega$ terms added to the $W_{QK}$ matrix. Most notably, pseudo-masking would interact poorly with most forms of dropout regularization and with $\ell^2$ regularization on the entries of $W_{QK}$.

\section{Discussion}

We have proven that attention heads can implement an MLP layer and in particular that any transformer can be converted to an attention-only transformer. One implication of these results is that it is theoretically possible to train an attention-only transformer that matches the performance of an MLP-plus-attention transformer. It remains unknown whether such an architecture would be competitive with the more classical transformer architecture in terms of practical considerations like training or inference speed. Such a test would be a promising future area of research. 

Our foremost hope in this work is to facilitate the advancement of mechanistic interpretability approaches such as \citet{elhage2021mathematical}, which found the most success in transformers without MLP layers, but found that a complete understanding of transformers ``will require progress on MLP layers''. Our technique could allow one to reuse the techniques that are successful on attention heads on the MLP layers.

In doing so, the primary impediment is scale since the approach described in this paper increases the number of attention heads in a transformer by several orders of magnitude. However, this is itself a useful new perspective on the difficulty of interpreting MLP layers: MLP layers in a model like GPT-3 are larger than attention layers by a 2:1 margin if one measures by number of parameters but by 500:1 if one measures by number of attention heads. It may be the case that the AI capabilities slogan ``scale is all you need'' applies equally to mechanistic interpretability.

\subsubsection*{Acknowledgements}

The authors would like to thank Ari Rahikkala for pointing us towards relevant literature and Delta Hessler for proofreading. The authors would like to thank Open Philanthropy for their support.

\bibliography{refs}
\bibliographystyle{iclr2024_conference}

\end{document}